\documentclass{article}

\usepackage{microtype}
\usepackage{graphicx}
\usepackage{subfigure}
\usepackage{booktabs} % for professional tables

\usepackage{hyperref}

\usepackage[accepted]{icml2024}

\usepackage{amsmath}
\usepackage{amssymb}
\usepackage{mathtools}
\usepackage{amsthm}

\def\bos{\text{BOS}}
\def\inp{\mathbf{x}}

\usepackage[capitalize,noabbrev]{cleveref}

\theoremstyle{plain}

\theoremstyle{definition}

\theoremstyle{remark}

\usepackage[textsize=tiny]{todonotes}

\icmltitlerunning{Tokenized SAEs: Disentangling SAE Reconstructions}

\begin{document}

\twocolumn[
\icmltitle{Tokenized SAEs: Disentangling SAE Reconstructions}



\icmlsetsymbol{equal}{*}

\begin{icmlauthorlist}
\icmlauthor{Thomas Dooms}{equal,independent}
\icmlauthor{Daniel Wilhelm}{equal,independent}
\end{icmlauthorlist}

\icmlaffiliation{independent}{Independent}
\icmlaffiliation{independent}{Independent}

\icmlcorrespondingauthor{Thomas Dooms}{doomsthomas@gmail.com}

\icmlkeywords{Sparse Auto-Encoders, N-grams, Class Imbalance}

\vskip 0.3in
]



\printAffiliationsAndNotice{\icmlEqualContribution} % otherwise use the standard text.

\begin{abstract}
Sparse auto-encoders (SAEs) have become a prevalent tool for interpreting language models' inner workings. However, it is unknown how tightly SAE features correspond to computationally important directions in the model. This work empirically shows that many RES-JB SAE features predominantly correspond to simple input statistics. We hypothesize this is caused by a large class imbalance in training data combined with a lack of complex error signals. To reduce this behavior, we propose a method that disentangles token reconstruction from feature reconstruction. This improvement is achieved by introducing a per-token bias, which provides an enhanced baseline for interesting reconstruction. As a result, significantly more interesting features and improved reconstruction in sparse regimes are learned.

\end{abstract}

\section{Introduction}

The holy grail of mechanistic interpretability research is the ability to decompose a network into a semantically meaningful set of variables and algorithms. SAEs have emerged as a promising method to extract interpretable context \citep{cunningham23, attention_saes, transcoders}. However, the importance of SAE features to model computation is still unknown. This paper specifically studies the importance of local context on the variety of learned features. This is enhanced by an SAE training token frequency imbalance resulting in bias toward local context.

We find that many features in medium-sized SAEs such as RES-JB \citep{neuronpedia} are affected by this imbalance. This causes them largely to reconstruct a direction biased toward the direction of the most prevalent training data unigrams. Empirically, we estimate that between 35\% and 45\% of the features reconstruct common unigrams and almost 70\% reconstruct common bigrams. We hypothesize these features then moreso reflect training token statistics than interesting internal model behavior. We attribute this phenomenon to the following two observations:

\begin{itemize}
    \item Local context is a strong approximation for latent representations, even in deeper layers.
    \item There is a prominent class imbalance in the training data of SAEs. Certain local combinations will appear much more frequently than specific global interactions.
\end{itemize}

Given both their frequency and strength in the representation, these local contexts occupy the majority of the features an SAE uses to minimize its reconstruction error. We show this to hold for all kinds of common $n$-grams. Furthermore, we hypothesize this to be the cause for a range of pathological behaviors exhibited by SAEs, such as the inability to generalize out-of-distribution in certain contexts \citep{templeton2024scaling, pathological_recons}.

Fortunately, these insights can be leveraged toward a solution; we propose a means to disentangle these "uninteresting" feature reconstruction tokens from the interesting features. This is accomplished by extending the SAE with a per-token bias,  allowing the SAE to represent a "base" reconstruction for each token. This leaves room for more semantically useful features. Furthermore, the proposed bias lookup table is efficient, resulting in SAEs becoming less compute-intensive to train. Specifically, our contributions are:
\begin{itemize}
    \item We identify and characterize the issue of SAEs learning token reconstruction features due to the input distribution and formulate why this is the case.
    \item We propose a technique to mitigate this behavior by separating token reconstruction from context reconstruction. We name this approach \emph{Tokenized SAEs}.
\end{itemize}

\section{Background}

\subsection{Notation} For interpretability, it is important to relate a sequence of $N$ input tokens $\inp \in \mathbb{T}^{N}$ to activations at some location $p$. This mapping exists as a function $A^p$.

We define an $n$-gram as $[t_0, t_1, t_2, \dots, t_n] \in \mathbb{T}^{n+1}$. In this paper, we assume $t_0 = \bos \in \mathbb{T}$, the beginning-of-sequence token.

\subsection{Imbalance} 
We will examine sparse auto-encoders at some location $p$. These map each row vector of $A^p(\inp)$ to itself, reconstructing it. The sparsity of the hidden layer is minimized, leading to seemingly interpretable features.

During training, short $n$-grams are exponentially over-represented due to an imbalanced training distribution. This biases the SAE toward reconstructing these short $n$-gram inputs.

This occurs because the SAE is trained to reconstruct each row vector of $A^p(\inp)$. Due to attention, each is a function only of the prior tokens, i.e. for row $i$, $A^p(\inp)_i = A^p(\inp_{\leq i})_i$. For example, for each training prompt the SAE is provided training examples $A^p(\bos)$ and $A^p(\bos, t)$, where $t$ follows the distribution of training set tokens.

\begin{figure}
    \centering
    \includegraphics[width=1.00\linewidth]{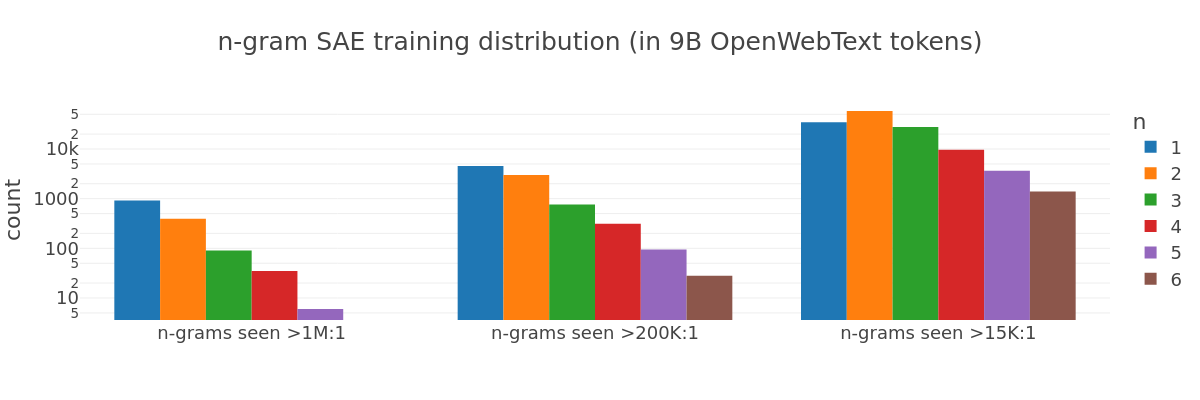}
    \caption{In the OpenWebText corpus, particular $n$-grams are seen exponentially more often than others. Many combinations occur millions of times more than an arbitrary $n$-gram.}
    \label{fig:n-grams}
\end{figure}

For row vector $i$, there are at most $|\mathbb{T}|^{i}$ possible activations. However, in practice the degree of over-representation can be measured directly for a given training set. Assuming each training sequence begins at a random token, the $n$-gram frequency distribution follows the dataset's $n$-token frequency distribution. We show many $n$-grams are more than a million times more likely than baseline (\autoref{fig:n-grams}) in the OpenWebText corpus \citep{openwebtext}.

This results in an effect similar to "imbalanced regression"\footnote{The terminology "imbalanced" accurately describes its implications, although it may best be described as a weighted class.}, where the target space distribution is sampled unevenly during training \cite{yang21, stocksieker24}. Each row vector follows an distribution based on its index, causing the SAE to become biased toward the highest-weighted regions of space (here the most common small $n$-gram activations). Such a class weighting causes a general MSE-trained regressor to underestimate rare labels \cite{jiawei22}. 

We show experimentally this causes higher reconstruction loss for less common unigrams, since they must "overcome" the biases (\autoref{fig:reconstruction}).

\section{Sparse Auto-Encoders}

\begin{figure}
    \centering
    \includegraphics[width=0.8\linewidth]{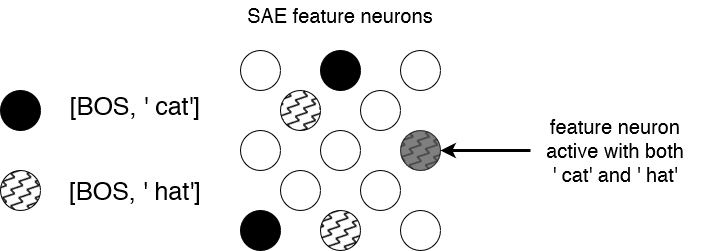}
    \caption{Illustrating experimental results, an individual feature neuron is activated when one of its associated $n$-grams is present. The most common tokens will occupy a full feature while less common tokens will share a feature. To maximize reconstruction, this sharing occurs between semantically similar tokens. }
    \label{fig:feature_neuron}
\end{figure}

The motivation for training SAEs is often presented as feature discovery. This is achieved by reconstructing the hidden representations through a sparse hidden basis, often called features. We show that SAEs memorize and organize themselves around the most common input $n$-grams, contributing to the observed correlation between them (\autoref{fig:feature_neuron}).

\subsection{Memorization} 
\begin{figure}
    \centering
    \includegraphics[width=1.00\linewidth]{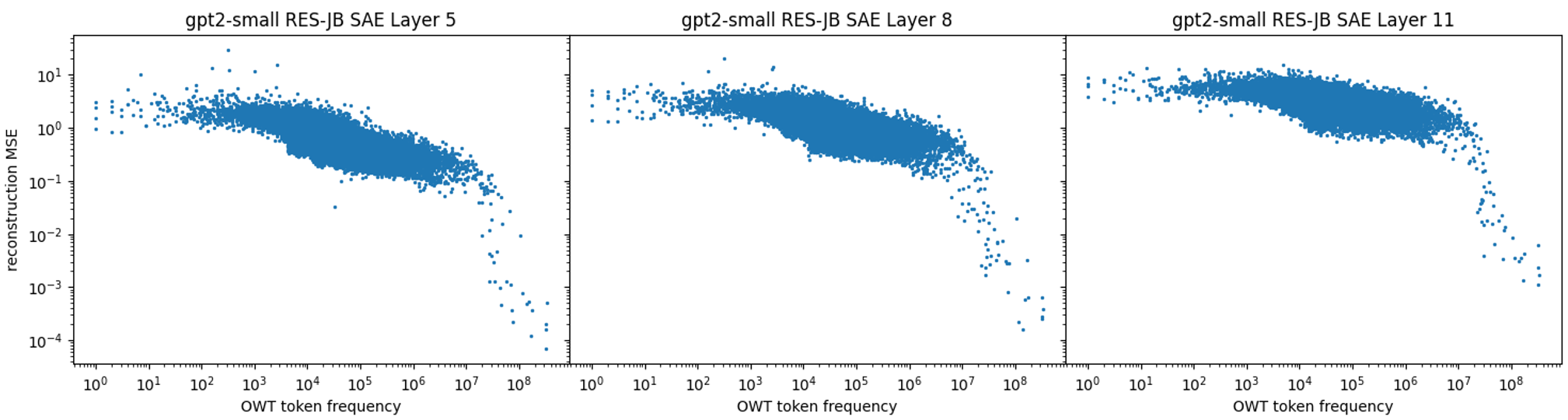}
    \caption{With increasing OpenWebText token frequency, the reconstruction MSE of unigrams in layers 5, 8, and 11 of the RES-JB SAE decreases. This indicates the SAE effectively memorizes the most common tokens. This effect is not as pronounced with bigrams, likely because they are composed of common unigrams and/or occupy unigram subspaces.}
    \label{fig:reconstruction}
\end{figure}

\begin{figure}
    \centering
    \includegraphics[width=0.8\linewidth]{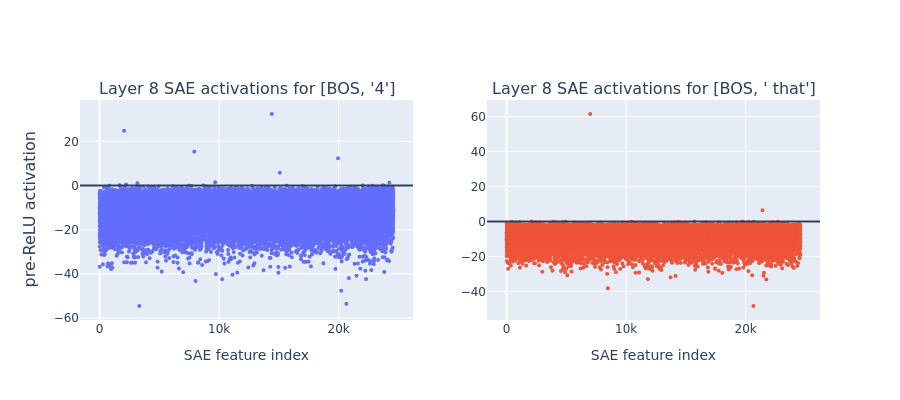}
    \caption{To memorize unigrams exactly and sparsely, the SAE represents each using a small subset of feature neurons that fire in response to the unigram. Due to the incorporation of prior token information, SAEs in later layers often also strongly memorize bigrams. }
    \label{fig:sparse_neurons}
\end{figure}

Suppose the most common $n$-gram inputs cause a training imbalance. Then we would expect to see (and observe) that with larger $n$-gram frequency, the reconstruction MSE decreases (\autoref{fig:reconstruction}) and fewer features activate (\autoref{fig:sparse_neurons}). In later layers, attention has likely consolidated information from other tokens, making the most common representations involve prior tokens. For example, many common words require multiple tokens to represent. We have observed evidence for this by noting that unigrams are most commonly activated in early layers and bigrams in later layers.

\subsection{Token Reconstruction Features} 
Suppose some SAE is represented by a set of features $\mathbb{F}$. Based on the prior experimental results and imbalance theory, we hypothesize that each common $n$-gram $\inp$ maps to a subset of $\mathbb{F}$ which $A^p(\inp)$ activates. The set of $n$-grams approaches a cover of $\mathbb{F}$, with the exclusion of dead features. (\autoref{fig:feature_neuron})

\begin{figure}[t]
    \centering
    \includegraphics[width=0.75\linewidth]{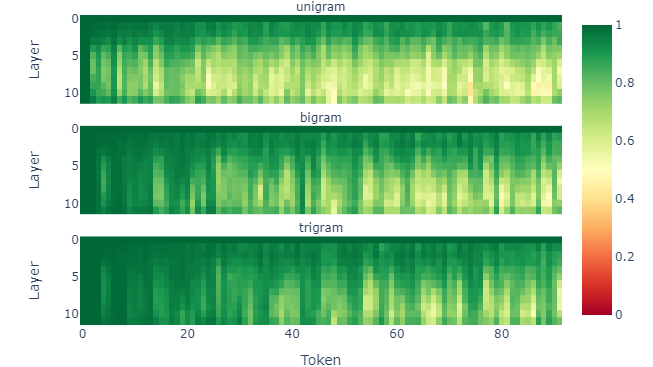}
    \caption{Measuring cosine similarity of hidden representations and a patched version which only includes the last $n$ tokens in GPT-2 small. Trigrams are generally an adequate approximation across the network.}
    \label{fig:patching}
\end{figure}

\begin{figure}
    \centering
    \includegraphics[width=0.9\linewidth]{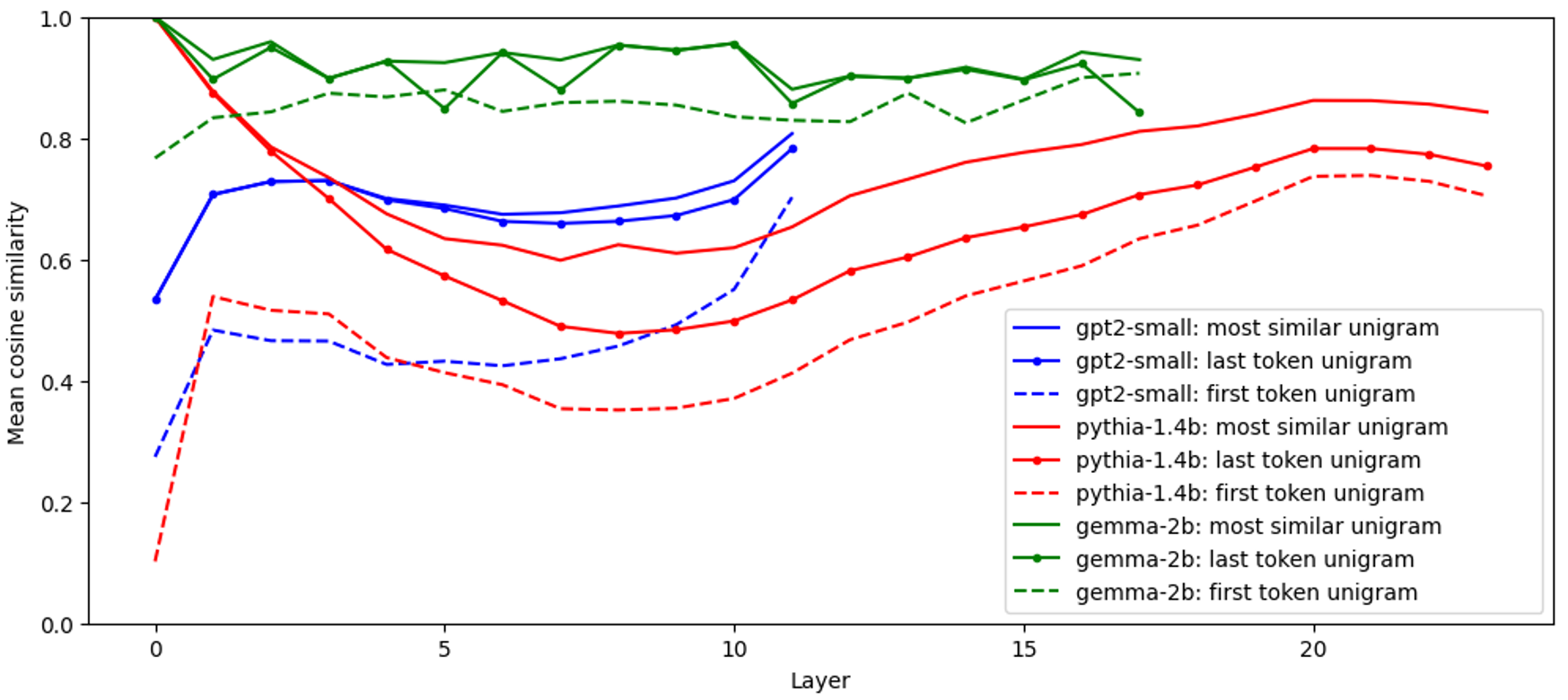}
    \caption{Tokenized SAEs assume residual activations are similar to those of their last-token  unigram. To study this, we compare the per-layer last-row residuals of 38K prompts to unigram residuals. We record their mean cosine similarity to (a) the most similar unigram, (b) the final-token unigram, and (c) the first-token unigram (as a control). We find that (a) residuals have very high cosine similarity to unigram residuals across all models and layers, and (b) the final token unigram is often nearly-closest.}
    \label{fig:complex_models}
\end{figure}

An SAE feature activates when a common activation pattern appears in $\inp$, corresponding to some aspect of an $n$-gram. We show this experimentally by predicting which input tokens will activate a given feature. In RES-JB layer 8, of the $76\%$ of features activated by a unigram, $39\%$ matched the top unigram activation and $66\%$ matched at least one. The $24\%$ of features not activated by a unigram illustrate:

\begin{enumerate}
    \item In later layers, some common SAE inputs may result from non-local information that more likely occurs in longer sequences. Experimental evidence shows that a minority of layer 8 GPT-2 features do not respond to any of 212K most-common ($n \leq 6$)-grams. A qualitative characterization of these features reveals these features exhibit more interesting semantic behavior.
    \item This method operates under the assumption that some $n$ tokens prior to row vector $i$ are sufficient to mostly describe the SAE inputs, i.e. $A^p(\inp)_i \approx A^p(\inp_{i-n})_i$. We show this to generally be the case in \autoref{fig:patching}, even in complex models and later layers (\autoref{fig:complex_models}). See \autoref{app:analysis} for additional discussion.
\end{enumerate}

\section{Tokenized SAEs} \label{sec:tokenized}
To resolve the abovementioned issues, we propose a new method that separates token reconstruction features from the dictionary. This is achieved by adding a separate path to the SAE which provides a base reconstruction of tokens. Concretely, we add a lookup table which acts as a per-token bias (\autoref{eq:lookup}).

\begin{align}
\mathbf{f}(\mathbf{a_t})  = & \text{ReLU}(W_{\text{enc}}(\mathbf{a_t} - b_{dec}) + b_{enc}) \\
\hat{\mathbf{a}}_\mathbf{t} = & W_{\text{dec}}\mathbf{f}(\mathbf{a_t}) + b_{dec} + W_{lookup}(\mathbf{t})  \label{eq:lookup}
\end{align}

This lookup table has no impact on the encoding, thus computing feature activations requires no change in setup. However, the lookup vector of the last token $\mathbf{t}$ is necessary for the reconstruction. We provide further details in \autoref{app:training}.

\subsection{Training} \label{sub:training}
As with the encoder, sensibly initializing the lookup table leads to large improvements in learning speed and final convergence. We do so with each token's activations excluding context; formulaically,  $W_{lookup}(\mathbf{t}) = A^p([\bos, \mathbf{t}])_1$.

Since the lookup table is essentially a hyper-sparse set of features, it is necessary to increase its learning rate to yield better reconstructions. A sensible approach is to multiply the learning rate by the approximate L0 of the SAE, this theoretically results in equal gradient updates. However, empirical results indicate using even higher learning rates is beneficial. More training-specific information can be found in \autoref{app:training}.

\subsection{Reconstruction} 
The experiments in this section are all performed on layer 8 of GPT-2 small. This is sufficiently deep in the model that we would expect complex behavior to have arisen. Furthermore, a breadth of public pre-trained SAEs can be used for comparison. We use the \emph{added} cross-entropy (\autoref{eq:ce_added}) to measure the impact on the model prediction and \emph{normalized} MSE (\autoref{eq:nmse}) to measure reconstruction.

\begin{align} \label{eq:ce_added}
CE_{added}(\inp) = \dfrac{CE_{patched}(\inp) - CE_{clean}(\inp)}{CE_{clean}(\inp)}
\end{align}

\begin{align}  \label{eq:nmse}
\text{NMSE}(\inp) = \dfrac{||\inp - \text{SAE}(\inp)||_2}{||\inp||_2}
\end{align}

\begin{figure}[t]
    \centering
    \includegraphics[width=1.00\linewidth]{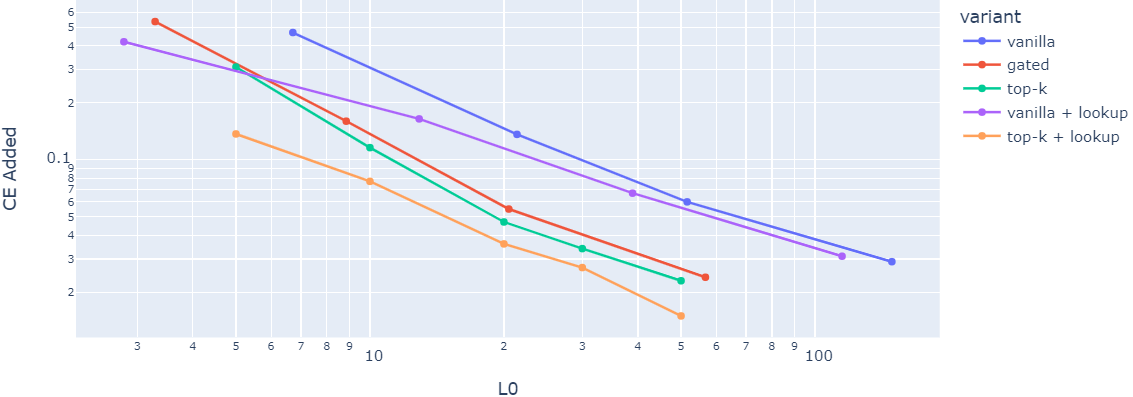}
    \includegraphics[width=1.00\linewidth]{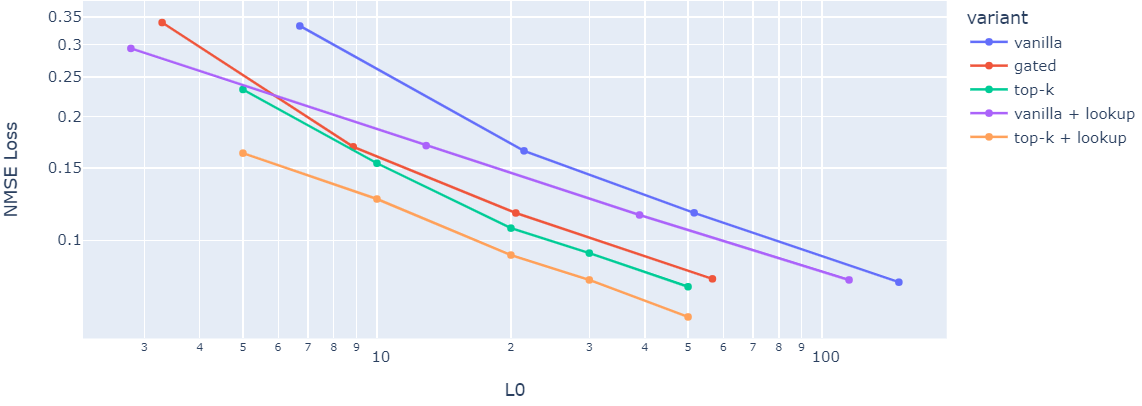}
    \caption{A Pareto frontier comparison of various SAEs on layer 8 of GPT-2 small using an expansion factor of 16. The cross-entropy added and normalized MSE compared to the L0 norm are shown. All SAEs were trained on about 300M tokens until (close to) convergence. Due to human error, the 'vanilla + lookup' did not learn its lookup table.}
    \label{fig:pareto}
\end{figure}

\autoref{fig:pareto} shows a large-scale comparison of Pareto frontiers for various architectures. We benchmark vanilla SAEs \citep{cunningham23}, gated SAEs \citep{gsae} and Top-k SAEs \citep{top-k}. The vanilla and the gated SAEs are trained with decoder sparsity loss from \citet{april_update}. Beyond this, no additional training techniques (resampling, ghost gradients, ...) were used.

This indicates Tokenized SAEs outperform their non-tokenized counterparts by a significant margin. They achieve the same reconstruction while being about 25\% sparser. Furthermore, in hyper-sparse regimes, they consistently yield good reconstructions and follow a consistently linear pattern compared to their counterparts.

\subsection{Suite}
\begin{figure}
    \centering
    \includegraphics[width=0.92\linewidth]{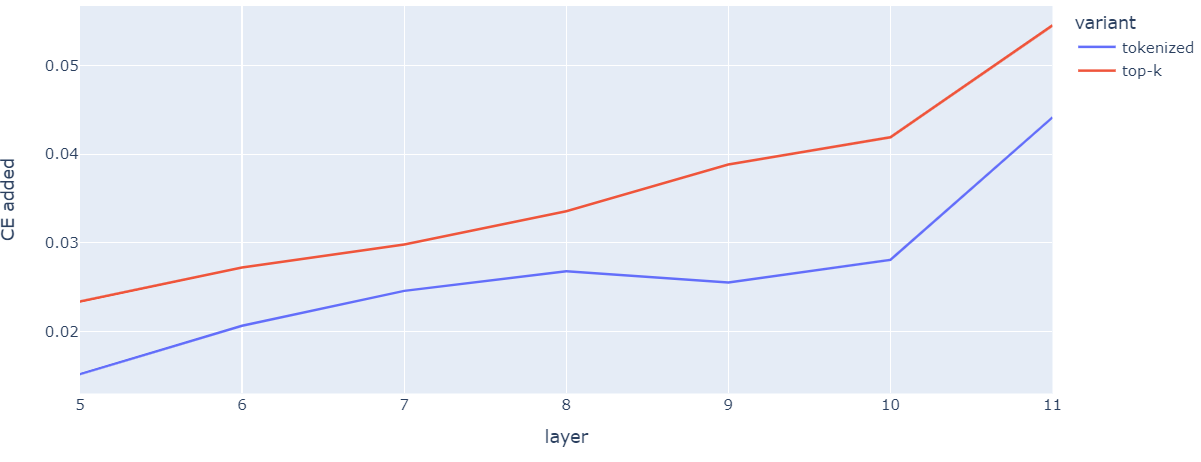}
    \includegraphics[width=0.92\linewidth]{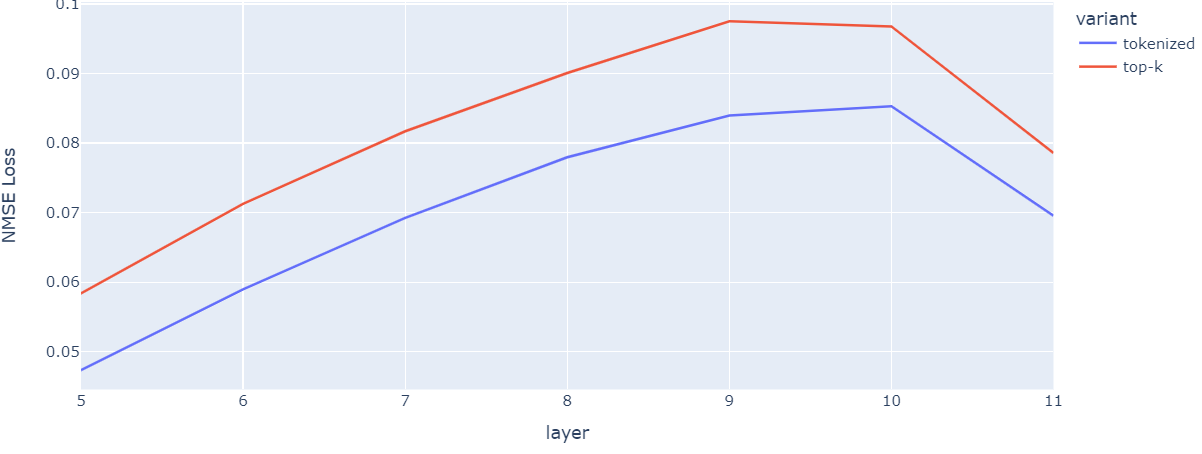}
    \caption{Comparison of a top-k SAE and tokenized variant on layers 5 through 11 of GPT-2 small with $k=30$. All SAEs were trained with an expansion factor of 16 with about 250M tokens.}
    \label{fig:layers}
\end{figure}

To demonstrate the generality of the approach, we train two suites of SAEs, a top-k and tokenized top-k, on layers 5 through 11 of GPT-2 small. These show that the lookup table consistently enhances reconstructions, with no visible degradation in deeper layers (\autoref{fig:layers}).

Additionally, TSAEs result in a significantly faster training speed. TSAEs reach the final value of baseline NMSE and CE added 6-10x faster across all layers of GPT-2. Training a competitive SAE (according to these metrics) can be achieved in mere minutes on consumer hardware.

\begin{table*}
\centering
\begin{tabular}{c|c|c|c|c|c|}
     & \textbf{RES-JB} & \textbf{Vanilla} & \textbf{Vanilla*} & \textbf{Top-k} & \textbf{Top-k*} \\
\hline
\textbf{Consistency}  & 4.1 & 3.6 & 3.4 & 3.4 & 4.2 \\
\textbf{Complexity}   & 2.5 & 1.1 & 2.9 & 1.7 & 3.0 \\
\end{tabular}
\caption{We manually score 20 features from multiple SAEs (tokenized denoted by an asterisk) and note their mean complexity and consistency according to \citet{june_update}. In short, the complexity score ranges from 1 (unigrams) to 5 (deep semantics). The consistency ranges from 1 (no discernable pattern) to 5 (no deviations). Given the limited sample size, these results should be interpreted cautiously; they provide preliminary indications rather than definitive evidence. Scoring features manually is time-consuming. }
\label{table:blind}
\end{table*}

\subsection{Scaling}
One salient concern for this approach is the impact of deeper and more complex models on the utility of the token lookup table. To this end, we perform preliminary experiments on Pythia 1.4B for layers 12, 16 and 20 using the newly proposed top-k. Results indicate that tokenized SAEs still outperform their baselines (\autoref{table:pythia}). This reinforces that token subspaces may be salient even in larger models.

\begin{table}[H]
\centering
\begin{tabular}{c|c|c|c}
  & \textbf{12}    & \textbf{16}    & \textbf{20}     \\
\hline
\textbf{Top-k}  & 0.076 & 0.081 & 0.155  \\
\textbf{Tokenized} & 0.045 & 0.055 & 0.121  \\
\end{tabular}
    \caption{CE added across 3 layers of Pythia-1.4B using top-k SAEs with $k=50$. Due to computation constraints the SAEs are undertrained, using only 70M tokens. Qualitatively, the training progression showed no signs of the baseline 'catching up'. The NMSE (not shown) exhibits a similar improvement.}
\label{table:pythia}
\end{table}

\section{Feature Comparison}

\subsection{Quantitative} 
\begin{figure}
    \centering
    \includegraphics[width=0.9\linewidth]{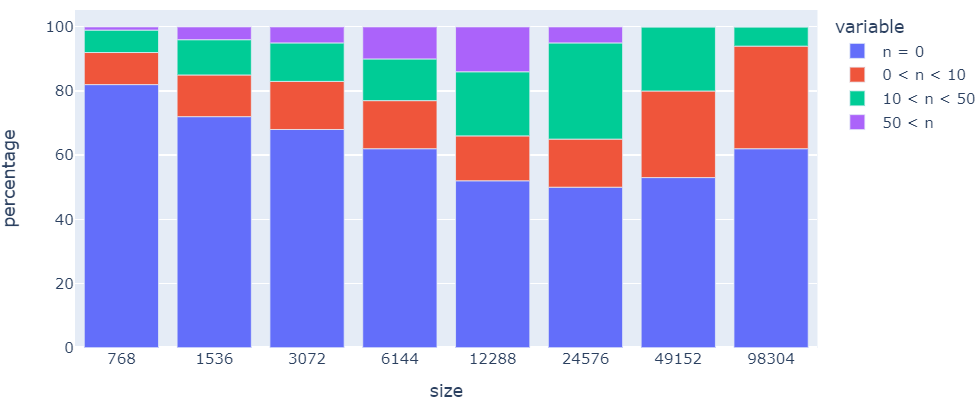}
    \caption{By SAE feature size \citep{jbsplit}, we show the percentage of features that are strongly activated (\textgreater5) by ranges of $n$ unigrams (with all categories summing to 100\%). In smaller SAEs, there is no bandwidth to represent individual or small sets of tokens. In medium-sized SAEs, we see features representing multiple tokens. As size increases further, features represent specific tokens.}
    \label{fig:tokens_active}
\end{figure}

We quantify the number of uninteresting features by forward-passing each possible unigram (prepended with $\bos$) and measuring the number of features that activate strongly for it. Features that strongly correspond to only a small set of tokens are more likely to be token reconstruction features, since they are at least partially responsible for reconstructing the tokens. We show that as SAE size increases, a smaller set of unigrams activates each feature (\autoref{fig:tokens_active}).

We perform the same experiment on the Tokenized SAEs from \autoref{fig:pareto}. We find that the number of features that activate on any single unigram is below 5\% for all of them. \autoref{app:analysis} contains more in-depth analyses regarding the differences.

\subsection{Qualitative} 
We performed a blind study on five GPT-2 layer 8 SAEs: a top-k and vanilla SAE, their tokenized counterparts, and RES-JB \citep{neuronpedia} as a baseline. The results are shown in \autoref{table:blind} and suggest that the TSAE features are about equally consistent, but their complexity is noticeably higher. \autoref{app:features} includes a list of cherry-picked features to corroborate these subjective findings. In summary, we find that features generated by Tokenized SAEs tend to be more semantically meaningful and contain fewer uninteresting features.

\section{Future Work}
Tokenized SAEs have a wide possible range of extensions. One candidate is to incorporate $n$-gram statistics, instead of simply unigrams. We believe this to be mostly an engineering challenge; it requires efficiently making a sparse, multi-token lookup table. Furthermore, while this paper only considers the tokens as a sparse basis, one could consider a previous SAE as a basis. This would incentivize structuring around already-existing features, likely improving circuit analysis.

Additionally, a more thorough study into the quality of Tokenized SAE features is still to be performed. This should be done on both the dictionary and the lookup table. The former is related to the incorporated non-local context and the latter is related to the token reconstruction. Exactly characterizing this token reconstruction similarity in latent representations is undoubtedly useful.

\section{Acknowledgements}
This project originated as a MATS sprint. We thank Jacob Dunefsky and Neel Nanda for their insightful discussions and guidance. We also thank Michael Pearce for coining the project's name. This research received funding from the Flemish Government under the "Onderzoeksprogramma Artificiële Intelligentie (AI) Vlaanderen” programme.

\section{Contributions}
Thomas conceived the proposed approach, trained the SAEs, and analysed the TSAEs and their differences. Daniel noticed and researched the training imbalance and analysed the TSAEs. The paper was written in tandem.


\bibliography{refs}
\bibliographystyle{icml2024}

%%%%%%%%%%%%%%%%%%%%%%%%%%%%%%%%%%%%%%%%%%%%%%%%%%%%%%%%%%%%%%%%%%%%%%%%%%%%%%%
%%%%%%%%%%%%%%%%%%%%%%%%%%%%%%%%%%%%%%%%%%%%%%%%%%%%%%%%%%%%%%%%%%%%%%%%%%%%%%%
% APPENDIX
%%%%%%%%%%%%%%%%%%%%%%%%%%%%%%%%%%%%%%%%%%%%%%%%%%%%%%%%%%%%%%%%%%%%%%%%%%%%%%%
%%%%%%%%%%%%%%%%%%%%%%%%%%%%%%%%%%%%%%%%%%%%%%%%%%%%%%%%%%%%%%%%%%%%%%%%%%%%%%%
\newpage
\appendix
\onecolumn

\section{Training Setup} \label{app:training}

\subsection{General} 
The setup is intentionally kept as simple as possible to avoid confounding factors. Specifically, no resampling or ghost gradients are used. All SAEs are trained on a subset of C4 tokens using a context length of 256. Tokens are collected in a buffer of size 128K and then sampled in batches of 4096 to train the SAE. We use the Adam optimizer with a learning rate of $1e^{-4}$ and a cosine annealing learning schedule. This base training setup is consistent across all experiments.

For all GPT-2 models, we use an expansion factor of 16 (12288 features). For the Pythia-1.4B, we use an expansion factor of 8 (16382 features).

\subsection{Initialization}
We initialize $W_{enc}$ with the response of the $W_{dec}$ for all SAEs. Further, as stated in \autoref{sub:training}, we initialize the lookup table to the activations for that token without tokens. Both these initialization procedures attempt to attain the same goal; approximate an identity. We therefore "balance" the lookup $W_{lookup}(tok) = \alpha A^p(\bos, tok)$ and the encoder $W_{enc} = (1-\alpha) W_{dec}^T$ using a hyperparameter that sums to one. While all values outperform non-tokenized SAEs, we found $\alpha=0.5$ to work well across all experiments.

Interestingly, we can approximate $\alpha$ during training using \autoref{eq:alpha} where $W_{original}$ is the lookup at the start of training. This reveals how much the SAE naturally steers towards learning balancing the lookup and SAE. We find that the middle layers of GPT-2 converge towards 0.6 and the later layers towards 0.5. The middle layers of Pythia-1.4B tens to 0.45 and the later ones to 0.4.

\begin{align}
\hat{\alpha} &= \dfrac{1}{n} \sum^n_i \dfrac{W_{original} \cdot W_{lookup}}{||W_{original}||^2_2}  \label{eq:alpha}
\end{align}

\subsection{Learning Rate}
In \autoref{sub:training}, we note that increasing the learning rate of the lookup table improves the reconstructions. The reasoning is that, due to the difference in sparsity, each entry in the lookup table is updated much less than the features in the SAE. We empirically find that increasing the learning rate to 0.01 (up to 100x higher than the global learning rate) yields good results. We attribute this to the lookup table being more stable to train and again to token frequency imbalance. One could also dynamically change the learning rate of each lookup entry based on this frequency, we did not try this.

\subsection{Memory and Compute Overhead}
Adding a lookup table does not impact training or inference time significantly since it is an extremely efficient operation. We noticed a 3-5\% increase in training time by introducing the table. In terms of memory overhead, the lookup table has a larger impact. For common SAE sizes, the memory requirements double. We do not think this to be an issue since SAEs are generally not memory-heavy; a whole GPT-2 suite would consume about 3GB of memory. If this is an issue, one could consider using a truncated lookup table, containing only the $n$ most common tokens.

\section{Cherry-Picked Features} \label{app:features}

\begin{figure}
    \centering
    \includegraphics[width=0.4\linewidth]{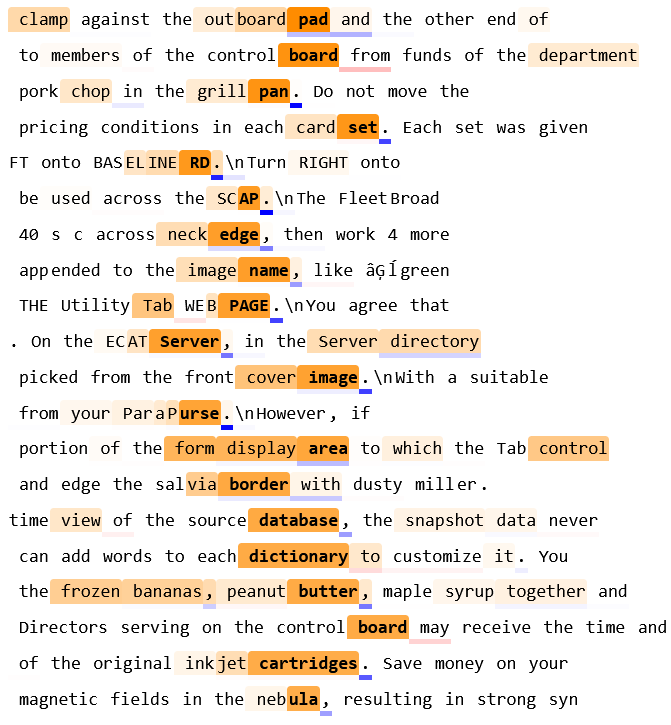}
    \caption{An end of sentence feature, boosting ".", ",", and "and" tokens.}
\end{figure}

\begin{figure}
    \centering
    \includegraphics[width=0.4\linewidth]{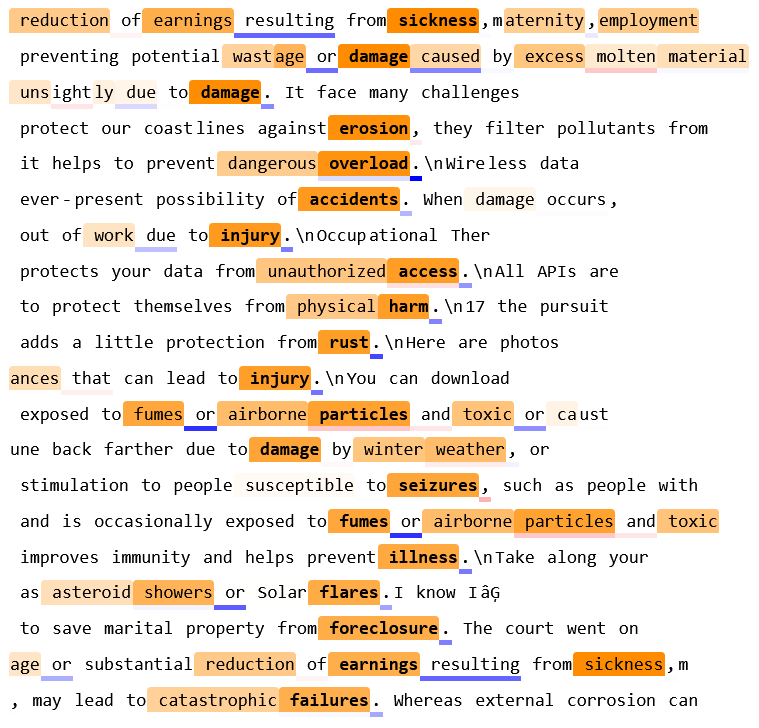}
    \caption{A health hazard feature.}
\end{figure}

\begin{figure}
    \centering
    \includegraphics[width=0.4\linewidth]{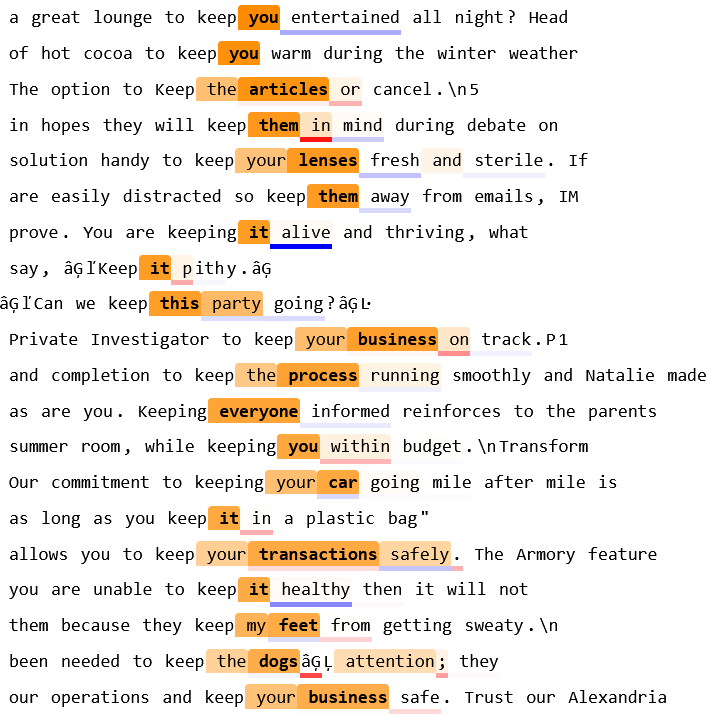}
    \caption{A direct object feature.}
\end{figure}

Potential Categories of First 25 Features (top-k TSAE, layer 8):
\begin{itemize}
\item \textbf{Overall thematic}: 16 (movie storylines)
\item \textbf{Part of a word}: 10 (second token), 12 (second token), 17 (single letter in a Polish word), 19 ("i/fi/ani")
\item \textbf{Thematic short n-grams}: 15 (" particular/Specific"), 23 (defense-related), 28 ("birth/death")
\item \textbf{N-grams requiring nearby period/newline/comma}: 7 ("[punctuation] If"), 18 ("U/u"), 22 ("is/be")
\item \textbf{Bigrams}: 2 ("site/venue"), 6 ("'s"), 8 ("shown that"/"found that"/"revealed that"), 14 ([punctuation] "A/An/a/ The")
\item \textbf{Categoric bigrams}: 13 ([NUM] "feet/foot/meters/degrees")
\item \textbf{Skipgrams}: 1 ("in the [TOK]"), 21 ("to [TOK] and")
\item \textbf{Locally Inductive}: 11 (requires a sequence of punctuation/short first names)
\item \textbf{Globally Inductive}: 24 (activates only when final token earlier in the prompt)
\item \textbf{Less Than 10 Activation (implies low encoder similarity with input)}: 0, 4, 5, 9
\item \textbf{Unknown}: 3, 20
\end{itemize}

Specific Interesting Features (top-k TSAE, layer 8):
\begin{itemize}
\item \textbf{36}: ".\textbackslash n[NUM].[NUM]"
\item \textbf{40}: Colon in the hour/minute "[1-12]:"
\item \textbf{1200}: ends in "([1-2 letters]"
\item \textbf{1662}: "out of [NUM]"/"[NUM] by [NUM]"/"[NUM] of [NUM]"/"Rated [NUM]"/"[NUM] in [NUM]"
\item \textbf{1635}: credit/banks (bigrams/trigrams)
\item \textbf{2167}: "Series/Class/Size/Stage/District/Year" [number/roman numerals/numeric text]
\item \textbf{2308}: punctuation/common tokens immediately following other punctuation
\item \textbf{3527}: [currency][number][optional comma][optional number].
\item \textbf{3673}: " board"/" Board"/" Commission"/" Council"
\item \textbf{5088}: full names of famous people, particularly politicians
\item \textbf{5552}: ends in "[proper noun(s)]([uppercase 1-2 letters][uppercase 1-2 letters]"
\item \textbf{6085}: ends in "([NUM])"
\item \textbf{6913}: Comma inside parentheses

\end{itemize}

Many features were found to activate on exact copies of the final $n$-gram. It is unknown if this is a possibility for all features.

\section{Additional Analysis} \label{app:analysis}

\subsection{Low feature activations imply low similarity with input vector}

\begin{figure}
    \centering
    \includegraphics[width=0.8\linewidth]{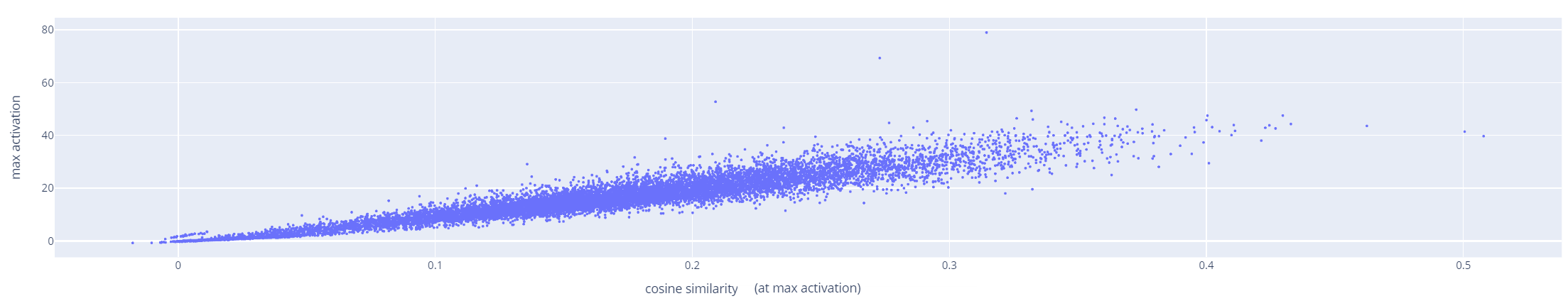}
    \caption{Due to how SAE activations are computed, feature activation strength is correlated with input vector cosine similarity with W\_enc. Low-activating features likely are not detecting signal in the input. (Figure shows top-k tokenized layer 8.)}
    \label{fig:tok8_cos_simil_correlation}
\end{figure}

It is important to ask whether an activated feature is detecting something of significance or not. One method to detect this is by the strength of the activation. The mechanics of the encoder computation indicate that larger feature activations will correlate with larger cosine similarity between the input vector and W\_enc (\autoref{fig:tok8_cos_simil_correlation}). Hence, a small-magnitude activation likely indicates the feature has not detected a signal.

Due to this, a minimum activation threshold is advisable when evaluating features.

\subsection{Recognizing dead features by encoder/decoder similarity}

\begin{figure}
    \centering
    \includegraphics[width=0.8\linewidth]{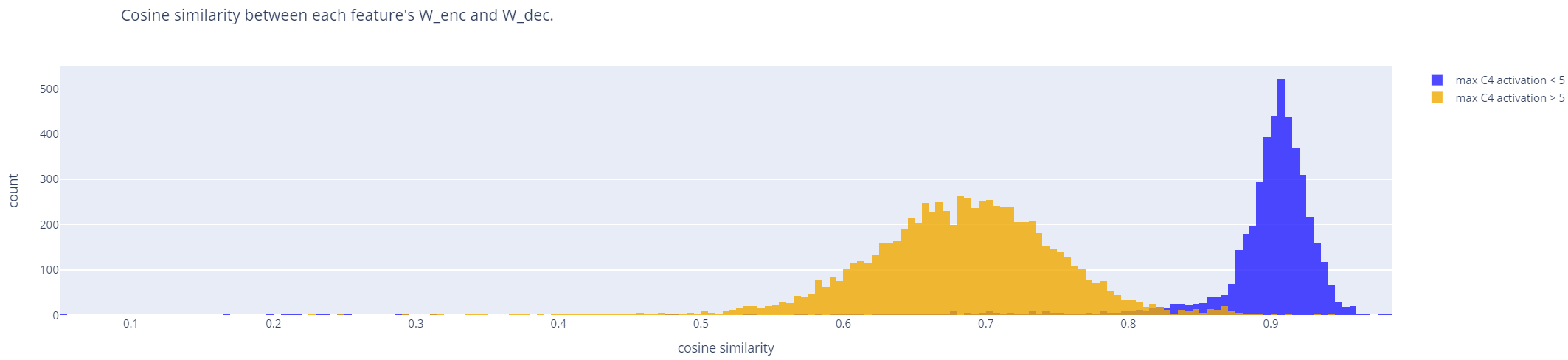}
    \caption{We initialize W\_enc as the transpose of W\_dec. This resulted in an easy post-facto test for dead features -- simply comparing the cosine similarity of the encoder and decoder. As shown, the peak at 0.9 exactly corresponds to features which never activate more highly than 3 (in our top-k tokenized SAE layer 8).}
    \label{fig:wenc_wdec_simil}
\end{figure}

Because we pre-initialize each feature with W\_enc and W\_dec transposed, an interesting finding is that dead features correspond nearly exactly to features with high cosine similarity between each feature's encoder and decoder. This can be used post-facto to detect dead features:

Dead features are evidenced by high cosine similarity between W\_enc and W\_dec, since they were pre-initialized as transposes (\autoref{fig:wenc_wdec_simil}). Here, we show these groups correspond nearly exactly to low test set activations (in gpt2-small layer 5 TSAE).

We examined the high-similarity group using four metrics, concluding they are likely not valid features:
\begin{itemize}
\item Nearly all are completely dissimilar to RES-JB features (\textless 0.2 max cosine similarity).
\item Nearly all have a top activation \textless 3 (activations are normally distributed about 0).
\item Nearly all are rarely (\textless1-10\%) in the the top 30 activations. (However, nearly all features with \textless 0.85 similarity are sometimes in the top 30.)
\item Manually looking at the activations, the features are often difficult to interpret.
\end{itemize}

\subsection{Feature complexity of TSAEs}

Measuring complexity is difficult, since feature activations may have multiple causes which are not yet fully understood. That said, a central motivation for TSAEs is that by excluding many "simple" unigram-based features, features may potentially represent more complex concepts (yet still be interpretable).

\begin{figure}
    \centering
    \includegraphics[width=0.8\linewidth]{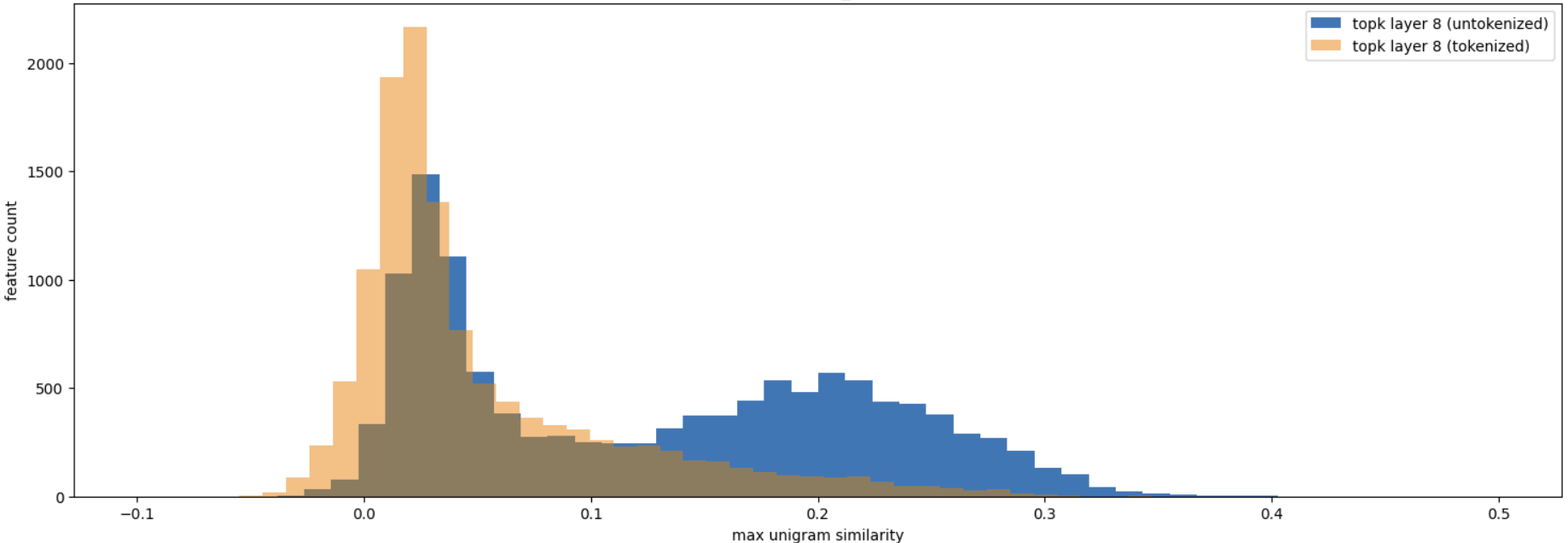}
    \caption{A central claim of TSAEs is that unigram-based features are reduced. Compared to a similarly-trained non-tokenized SAE, we see that significantly fewer TSAE features have significant max cosine similarity between the encoder weights and all unigram inputs. Hence, TSAE features will not respond as often to individual tokens.}
    \label{fig:unigram_comparison}
\end{figure}

\begin{figure}
    \centering
    \includegraphics[width=0.8\linewidth]{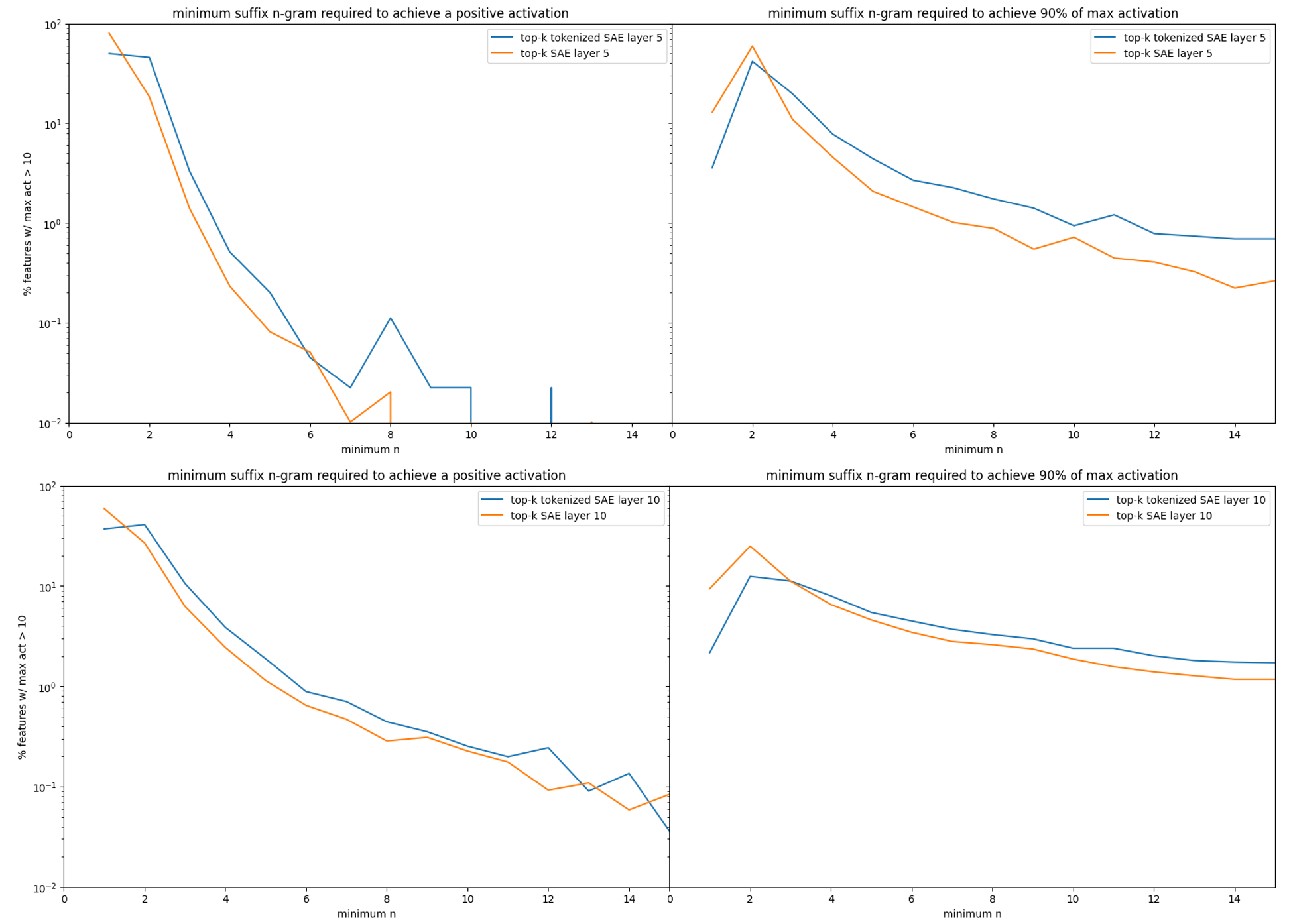}
    \caption{We measure feature complexity by finding the minimum suffix $n$-gram of each top-activating feature (\textgreater 10) that results in a positive activation (left) or 90\%-max activation (right). We note that a larger percentage of features in non-tokenized SAEs are unigrams ($n = 1$), while for $n > 2$ TSAEs generally have more "complex" features by this metric. Further, we see that layer 5 (top) achieves positive activations entirely with small $n$ compared to layer 10 (bottom).}
    \label{fig:min_ngrams}
\end{figure}

\begin{itemize}
\item First, we show that TSAE features are largely no longer unigram-based when compared to an identically trained non-tokenized top-k SAE. To measure this, we determine the max cosine similarity between all unigram input vectors and each feature's encoder weights. We find that W\_enc is drastically less similar to unigram features in a tokenized SAE (\autoref{fig:unigram_comparison}).

\item Second, we determine whether the additional features may be considered more "complex". To measure this, we examine features only with a minimum max activation to ensure they are not dead and properly detect some signal. Taking the top-activation prompt, we activate increasingly large suffix $n$-grams until the activation becomes (a) positive and (b) within 90\% of the maximum activation (to avoid outlier maximum indices). The former often indicates the beginning of an increasing activation, while the latter indicates a strong encoder weight similarity to the input. 

Plotting the percentage of features at each minimum $n$ (\autoref{fig:min_ngrams}), we notice that indeed TSAEs have more features activating at each $n > 2$ than a similarly-trained non-tokenized SAE. At least for this metric, we conclude that indeed the loss of unigram features translated into additional features requiring longer context.

\end{itemize}

\subsection{More on final-token subspaces}

\begin{figure}
    \centering
    \includegraphics[width=0.8\linewidth]{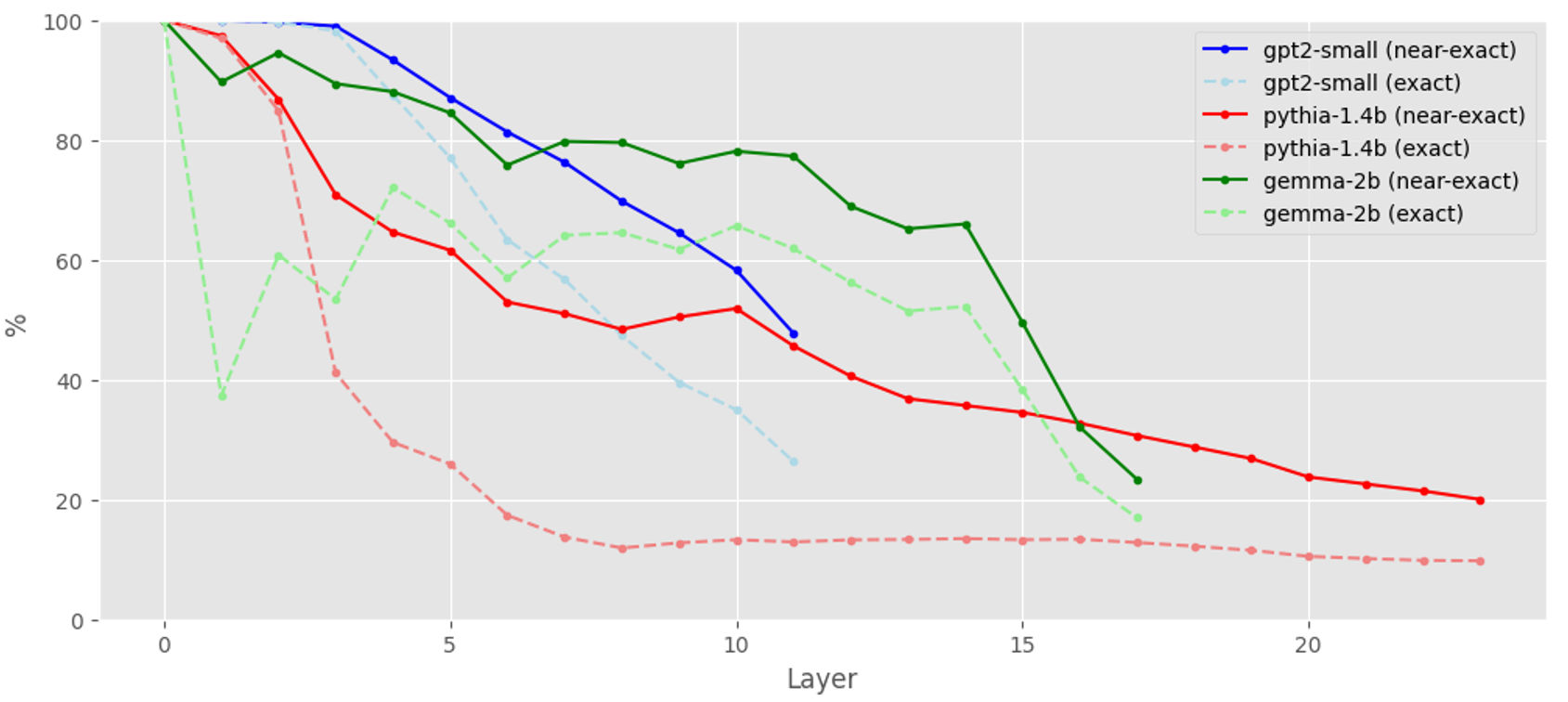}
    \caption{At each layer, we measure how many example prompts pre-layer residual activations are nearest to their last-token unigram activation than any other. Even in the final layer a large percentage are closest. When making a near-exact comparison, we compare token strings after stripping whitespace and lowercasing. (Typically, these unigram activations are nearby in space.)}
    \label{fig:pct_final_token_closest}
\end{figure}

Here, we provide additional support that resid\_pre activations are strongly related to a token subspace. We find that regardless of model complexity and layer -- and even with Gemma 2B's 256K vocabulary -- \textgreater 20\% of the time a prompt's final-token (or a near-exact token) residual is closer than any other unigram residual (\autoref{fig:pct_final_token_closest}).

\subsection{Bigram MSE reconstructions}

\begin{figure}
    \centering
    \includegraphics[width=1.0\linewidth]{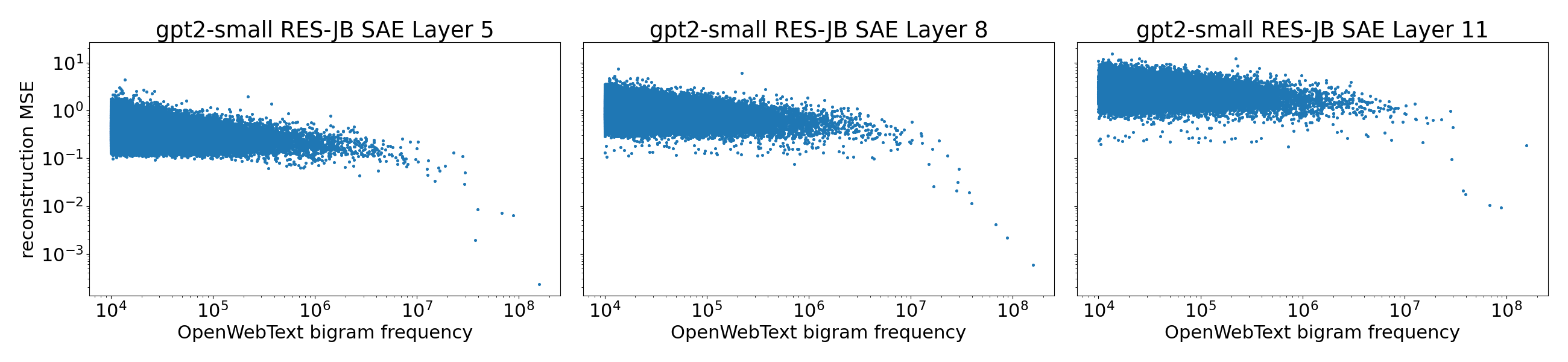}
    \caption{The 92K most common bigrams in OpenWebText also exhibit lower reconstruction MSE with larger training set frequency.}
    \label{fig:reconstruction_mse_bigrams}
\end{figure}

We find that for RES-JB SAEs, a larger bigram training set frequency also results in lower reconstruction MSE. However, since fewer bigrams occur as frequently and since they are made up of the most common unigrams, there is not as strong an effect. Some of the most common bigrams that exhibit a lower MSE include "\textbackslash{n}\textbackslash{n}", ". \textbackslash{n}", " of the", " in the", ", and", ", the", ". The", "\textbackslash{n}The", ", but", " on the", as well as multi-byte Unicode representations for single- and double-quotation marks. (\autoref{fig:reconstruction_mse_bigrams}).

\section{Neuronpedia Feature Study}

\begin{table}[h!]
\centering
\begin{tabular}{|c|l|l|}
\hline
\textbf{Index} & \textbf{Term} & \textbf{Type} \\
\hline
0 & numbers & Unigram collection \\
\hline
1 & “Pier” & \textbf{Unigram} \\
\hline
2 & weeks/months/years & Unigram collection \\
\hline
3 & Token after sorry/apologize & Bigram collection \\
\hline
4 & separator/time & Attention \\
\hline
5 & “in” & \textbf{Unigram} \\
\hline
6 & Adjectives related to famousness & Unigram collection + attention \\
\hline
7 & "recipe"/"recipes" & \textbf{Unigram} \\
\hline
8 & Causality (by a/due to) & Bigrams + attention \\
\hline
9 & “Ļ” & \textbf{Unigram} \\
\hline
10 & “Ļ” (again, look it up) & \textbf{Unigram} \\
\hline
11 & Not sure & Attention \\
\hline
12 & “told” & \textbf{Unigram} \\
\hline
13 & solved, addressed, resolved & Unigram collection \\
\hline
14 & “example” & \textbf{Unigram} \\
\hline
15 & Really not sure… & \textit{nan} \\
\hline
16 & “With” & \textbf{Unigram} \\
\hline
17 & “Ste” & \textbf{Unigram} \\
\hline
18 & numerics in brackets (references) & Bigram collection \\
\hline
19 & “s” after number (20s) & Bigram collection \\
\hline
20 & Anglo + Alred + Pf & Unigram collection \\
\hline
\end{tabular}
\caption{A qualitative study into the first 21 features of Joseph Blooms GPT-2 resid\_pre SAE on layer 8. We show that more than half of the features represent uninteresting reconstructions.}
\label{table:1}
\end{table}

\end{document}